\documentclass[authoryear]{elsarticle}
\usepackage{amsfonts}
\usepackage{lineno,hyperref}
\usepackage{amsfonts}
\usepackage{amsmath,amsthm}
\usepackage{amsmath}
\usepackage{amssymb}
\usepackage{amsfonts}
\usepackage{multirow}
\usepackage{amsthm}
\usepackage{caption}
\usepackage[table,xcdraw]{xcolor}
\DeclareMathOperator{\di}{d\!}
\usepackage{graphicx,psfrag,epsf,float}
\usepackage{enumerate}

\usepackage{natbib}
\setcitestyle{authoryear,open={(},close={)}}
\hypersetup{colorlinks,citecolor=red, filecolor=black, linkcolor=blue, urlcolor=black} 
\usepackage{url} 
\theoremstyle{plain}

\def\spacingset#1{\renewcommand{\baselinestretch}%
{#1}\small\normalsize} \spacingset{1}

\biboptions{authoryear}
 
\begin{document}

\begin{frontmatter}

\title{Neural Networks as Functional Classifiers}

\author[mymainaddress]{Barinder Thind}
\author[mymainaddress2]{Kevin K. S. Multani}
\author[mymainaddress]{Jiguo Cao\corref{mycorrespondingauthor}}

\address[mymainaddress]{Department of Statistics and Actuarial Science,\\ Simon Fraser University}
\address[mymainaddress2]{Department of Physics,\\ Stanford University}


\cortext[mycorrespondingauthor]{Corresponding Author. Postal Address: 8888 University Dr, Burnaby, BC, Canada, V5A1S6. Tel:(+1)778-782-7600; Fax: (+1)778-782-4368; Email: jiguo\_cao@sfu.ca}

\begin{abstract}
In recent years, there has been considerable innovation in the world of predictive methodologies. This is evident by the relative domination of machine learning approaches in various classification competitions. While these algorithms have excelled at multivariate problems, they have remained dormant in the realm of functional data analysis. We develop notable deep learning methodologies to the domain of functional data for the purpose of classification problems. We highlight the effectiveness of our method in a number of classification applications such as classification of spectrographic data. Moreover, we demonstrate the performance of our classifier through simulation studies in which we compare our approach to the functional linear model and other conventional classification methods.
\end{abstract}


\end{frontmatter}



\newpage
\spacingset{1.45} 

\section{Introduction}

In the zoo of statistical problems, classification has not ceased to be a central issue across numerous disciplines. Often, many tasks can be re-stated to be classification in nature and so, there is great benefit in developing methods that provide predictive and interpretive solutions to such problems. 
%
This article presents one such tool in the form of a functional expansion to deep learning methodologies, namely neural networks, with an exclusive focus on classification.

Given that this is such a central task in statistics, there is no shortage of multivariate approaches for classification. The canonical example is that of logistic regression, the general form of which is:
\begin{align*}
   P(y^{(i)}=1)=\frac{1}{1+\exp(-(\beta_{0}+\beta_{1}x^{(i)}_{1}+\ldots+\beta_{p}x^{(i)}_{p}))},
\end{align*}
where the set of parameters $\boldsymbol{\beta}$ are estimated through least squares methods and $y^{(i)} \in \{0, 1\}$ is the response variable. This can easily be extended for multiple classes in the form of multinomial regression. Tree methods such as random forests \citep{breiman2001random} and gradient boosted trees \citep{friedman2001greedy} have also been shown to have high accuracy in classification problems. Another approach, the support vector machine \citep{cortes1995support}, bins observations into classes by creating a hyperplane in a high dimensional space. This can be done with any of a number of kernels. Many alternatives involving support vector machines have also been proposed \citep{suykens1999least}. Another common approach is that of $k$-nearest neighbors \citep{peterson2009k}. This is a non-parametric approach that takes into account observations that are close to one another in space. Finally, we can consider neural networks \citep{mcculloch1943logical} which are made of neurons and layers. The general form of a given layer is: 
\begin{align*}
\boldsymbol{v}^{(1)} = g\left(\boldsymbol{W}^{(1)}\boldsymbol{x} + \boldsymbol{b}^{(1)}\right),
\end{align*}
where $\boldsymbol{x}$ is a vector of $J$ covariates, $\boldsymbol{W}^{(1)}$ is an $n_1$ x $J$ weight matrix, $\boldsymbol{b}^{(1)}$ is the intercept, and $g(\cdot)$ is some activation function that transforms the resulting linear combination \citep{ESL}. This family of methods has proven to be an excellent classifier and is at the bedrock of the methodology presented in this article.

With respect to functional data analysis, there does indeed currently exist a family of functional methods that try their hand at classification. For example, we can build up to the generalized functional linear model by first considering the conventional functional linear model. Let $Y$ be a scalar response variable, and $X(t)$, $t \in \mathcal{T}$, be the functional covariate. When the scalar response $Y$ follows a normal distribution, this model is defined as:
\begin{align*}
    E(Y|X) = \alpha + \int_{\mathcal{T}}\beta(t)X(t)\di t,
\end{align*}
where $\alpha$ is the intercept term, and $\beta(t)$ is the functional coefficient which represents the overall linear effect of $X(t)$ on $Y$ \citep{cardot1999functional}. The expansion to the general linear model is trivial to the multivariate counterpart and is written as: $E(Y|X) = g\left(\alpha + \int_{\mathcal T}\beta(t)X(t)\di t\right)$. In this case, the scalar response $Y$ follows a general distribution in an exponential family, $g(\cdot)$ is referred to as the link function and, in the case of a two class problem, is called the logistic function \citep{muller2005generalized}. Another extension, referred to as the functional partial least squares method, was proposed by \cite{preda2007pls}. This was further innovated on by \cite{aneiros2006semi} in the way of the semi-functional partial linear model defined as:
\begin{align*}
    E(Y|X) = r\left(X(t)\right) + \sum_{j=1}^J w_{j}Z_{j},
\end{align*}
where $Z_j, j = 1,\ldots,J,$ is the set of scalar covariates you would expect to see in the multivariate counterpart, and the function $r(\cdot)$ is also estimated with no parametric form using kernel methods.  The functional non-parametric approach was introduced by \cite{ferraty2006nonparametric} and took the form
\begin{align*}
    E(Y|X) = r\left(X(t)\right),
\end{align*}
where $r(\cdot)$ is a smooth non-parametric function that is estimated using a kernel approach. These models have their own merits however, for the classification examples given in this paper, we showcase the relative prowess of the methods introduced in this paper. Our proposed methodology takes the form:
\begin{align}
\label{eq1}
    v = g\left(\sum_{k = 1}^{K}\int_{\mathcal T} \beta_{k}(t)x_{k}(t)\di t + \sum_{j = 1}^{J}w_{j}z_{j} + b\right),
\end{align}
where $g(\cdot)$ is some activation (i.e. non-linear) function, $w_{j}$ is the weight associated with the scalar covariate $z_{j}$, and $\beta_{k}(t)$ is the functional weight that corresponds to the functional covariate $x_k(t)$. The formulation of \hyperref[eq1]{Equation (1)} is specifically in reference to a single neuron with $K$ functional covariates and $J$ scalar covariates. The final output of this general neural network will be a matrix containing the class probabilities for each of the input observations. 

There are a number of reasons to motivate this marriage of functional data analysis and neural networks. While \cite{rossi2005functional} introduced a neural network that showcased neural networks with a functional covariate, there has been significant innovation since then that significantly improves performance. For instance, residual neural networks, as presented in \cite{he2016deep}, overcame the problem of the vanishing gradient by adding an additional term to the set of neurons in a given layer. This innovation allowed users to implement models with hundreds of layers. There also was the introduction of several modern optimization techniques \citep{kingma2014adam, ruder2016overview}. These methods resulted in more efficient paths to the optima as well as a more accurate set of estimated parameters as measured by various error functions. When speaking about these innovations, we also must mention the plethora of hyperparameters that have decreased error rates across the board. Early stopping \citep{yao2007early} and dropout \citep{srivastava2014dropout} are two such hyperparameters. The former allowed users to halt the learning process early under the condition that there was not a significant decrease in error for some number of iterations. The latter randomly dropped neurons between layers; this prevented overfitting.

This paper has two key contributions. The first is a methodology that fuses neural networks with the general functional data. We present the methodology, the math that underpins the approach, as well as information about the plethora of hyperparameters. Second, we introduce an R package that can be generally used as a machine learning tool for longitudinal data. This package, \textbf{FuncNN} is readily available for use and can be downloaded from the Comprehensive R Archive Network \citep{thind2020funcnn}. We also note that, while not explored in-depth here, a by-product of the fusion is a potentially interpretive visualization of the network parameters that is not possible in the conventional neural network setting.

We explore the introduced method through application in a number of problems. For example, we consider a comparison between a number of functional methods for a fungi data set where we attempt to use functional covariates to classify into one of 18 classes. We also present multiple examples comparing this method to both functional and multivariate methods such as the classification of orange juice samples with a particular level of sugar contents. We also showcase the prowess of the approach through simulation studies.

The rest of the article is organized as follows. In \hyperref[sec:2]{Section 2}, we reiterate the neural network methodology and then extend it to be the functional neural networks alluded to earlier. This section also contains information regarding functional weights and the various hyperparameters of the network. In \hyperref[sec:3]{Section 3}, we apply the functional neural network on real-world data and highlight the results. Next, \hyperref[sec:4]{Section 4} contains simulation studies that further solidify the usefulness of the methods introduced here. This includes prediction comparisons of multivariate and functional methods in various simulation scenarios. Lastly, \hyperref[sec:5]{Section 5} contains conclusions and future discussions.

\section{Methodology}\label{sec:2}
We begin with an overview of neural networks. Let $n_{u}$ be the number of neurons in the $u$-th hidden layer. Using the previously defined notation, the general form of a layer in a neural network takes the form of:
\begin{align*}
    \boldsymbol{v}^{(1)} = g\left(\boldsymbol{W}^{(1)}\boldsymbol{x} + \boldsymbol{b}^{(1)}\right),
\end{align*}
where again,  $\boldsymbol{x}$ is a vector of $J$ covariates, $\boldsymbol{W}^{(1)}$ is an $n_1$ x $J$ weight matrix, $\boldsymbol{b}^{(1)}$ is the intercept (or the bias), and $g(\cdot)$ serves to perform some non-linear transformation of the linear combination input. The choice of the activation function $g:\mathbb{R}^{n_1} \rightarrow \mathbb{R}^{n_1} $ is largely dependent on the context of the problem -- although the ``relu'' \citep{hahnloser2000digital} and sigmoidal \citep{han1995influence} activation functions are commonly used. The output vector $\boldsymbol{v^{(1)}}$ is passed on to the next layer where a similar functional transformation takes place -- this process is repeated until the final layer where usually, the softmax function \citep{bishop2006pattern} will output class probabilities. The learning process involves incremental updates of the weights and biases of the network by one of the plethora of gradient descent algorithms available -- details are provided later in the article.

While this model structure is perfectly compatible with the traditional multivariate format of data, it falls short when we try and extend into the domain of functional data analysis (FDA) \citep{ramsay2009}. This shortcoming begins with the underlying assumption about the input object, $\boldsymbol{x}$. The fundamental difference between FDA and conventional multivariate methods is the paradigm through which data is observed. In FDA, data takes a functional form: $x(t): t \rightarrow \mathbb{R}$. For the purpose of intuition, consider observations of some realization of a phenomenon over its corresponding continuum; traditionally, we would treat each individual observation as a discrete point however, in the case of FDA, we consider instead a single curve that represents the behaviour over the continuum. In order to estimate this functional observation, basis expansion methods are used, with Fourier and B-spline methods being particularly favorable. The general form of these expansions is $x(t) = \sum_{i = 1}^{n} c_{i}\phi_{i}(t)$, where $\boldsymbol{\phi(t)}$ is the set of basis functions corresponding to a particular expansion and $\boldsymbol{c}$ is the vector of coefficients. These coefficients are the particulars that separate between realizations and can be estimated in a number of ways \citep{Ramsay05}.

We now consider the neural network expansion for the case when our input is infinite dimensional defined over some finite domain $\mathcal{T}$, i.e., our input is a functional covariate $x(t): \mathcal{T} \rightarrow \mathbb{R}$, $t\in\mathcal{T}$. In the traditional neural network, the weight matrix $\boldsymbol{W}$ puts a weight on each of the $J$ input observations -- since our input observation is infinite-dimensional, our weight must also be infinite dimensional i.e. a function. This weight is defined as $\beta(t)$. The form of a neuron can then be observed as:
\begin{align*}
    v^{(1)}_{i} = g\left(\int_{\mathcal T}\beta_{i}(t)x(t)\di t + b_{i}^{(1)}\right),
\end{align*}
where the subscript $i$ is an index that denotes one of the $n_{1}$ neurons in this first hidden layer, i.e., $i\in \lbrace{1,2,\dots,n_1\rbrace}$. Similar to the expansion for $x(t)$, we can define the functional weight $\beta_i(t)$ as an M-term linear combination of basis functions: $\beta_i(t) = \boldsymbol{c}_i^T \boldsymbol{\phi}_i(t)$, where $\boldsymbol{\phi}_i(t) = (\phi_{i1}(t), \ldots, \phi_{iM}(t))^T$ is a vector of basis functions, and $\boldsymbol{c}_i = (c_{i1}, \ldots, c_{iM})^T$ is the corresponding vector of basis coefficients. A key difference between the estimation of $\beta(t)$ and $x(t)$ is that the coefficients for the functional weight will be initialized randomly and updated in an iterative fashion through the neural network learning process. Since this is only the format of the first layer, the rest of the $u - 1$ layers of the network can be of any of the usual forms such as feed-forward or residual layers. Using these basis approximations of $\beta_i(t)$, we can simplify to get that the form of a single neuron is:
\begin{align}
\label{eq:single_neuron}
    v_{i}^{(1)} &= g\left(\int_{\mathcal T} \beta_{i}(t)x(t)\di t + b_{i}^{(1)}\right)\nonumber\\
    &= g\left(\int_{\mathcal T} \sum_{m=1}^{M} c_{im}\phi_{im}(t)x(t)\di t + b_{i}^{(1)}\right)\nonumber\\
    &= g\left(\sum_{m=1}^{M} c_{im}\int_{\mathcal T}\phi_{im}(t)x(t)\di t + b_{i}^{(1)}\right),
 \end{align}
\noindent where the integral in \hyperref[eq:single_neuron]{Equation~(2)} can be approximated with numerical integration methods such as the composite Simpson's rule \citep{Suli03}.

\begin{figure}[h!]
  \centering
  \includegraphics[scale = 0.4]{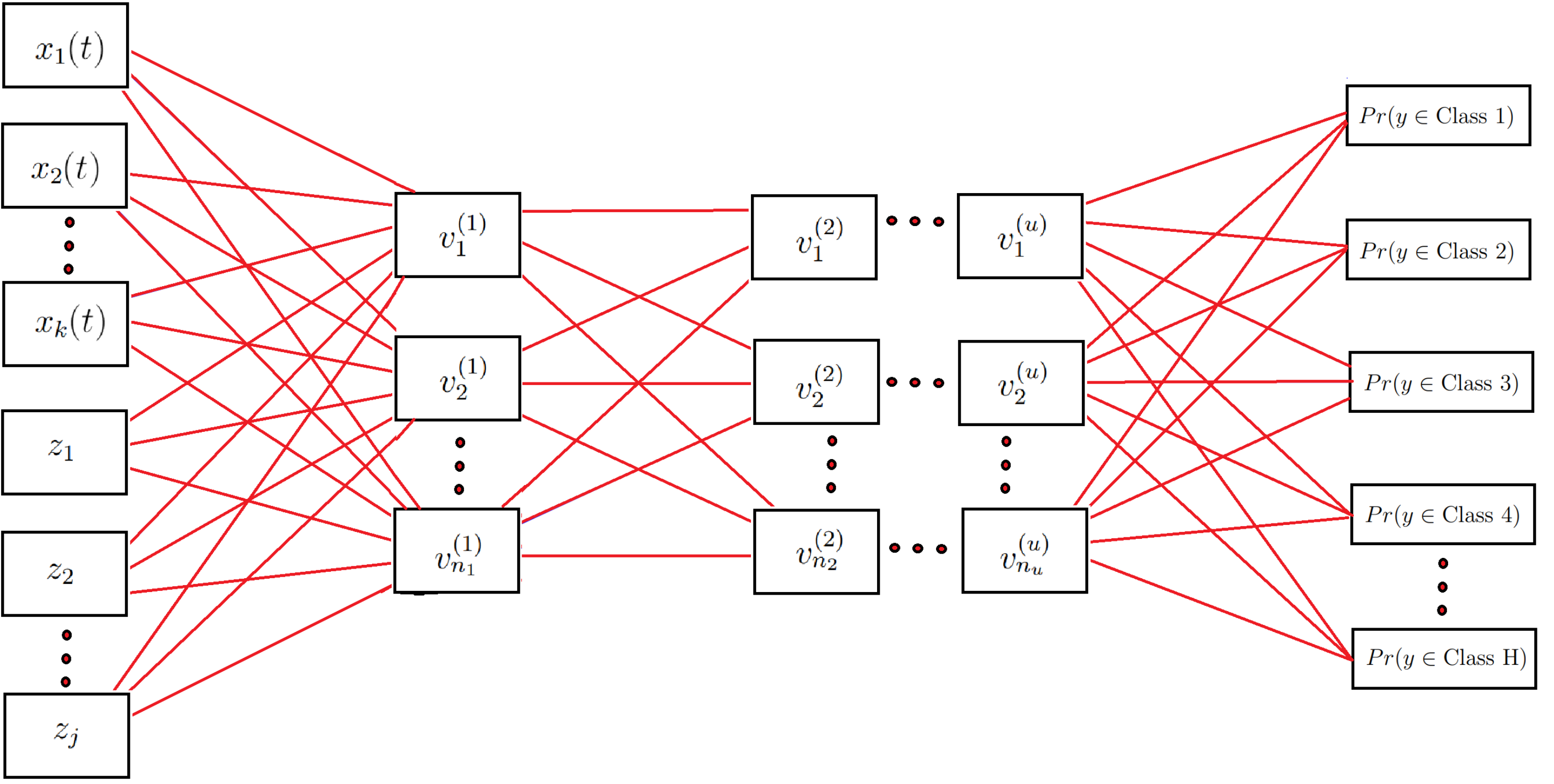}
  \label{fig:1}
  \caption{Schematic of a general functional neural network for when the inputs are functions, $x_k(t)$, and scalar values, $z_j$. The response/output of this network is a vector of probabilities where each element corresponds to one of the $H$ classes.}
\end{figure}

As is in the case of the traditional data, we can observe multiple functional covariates for a single observation. The generalization for $K$ functional covariates and $J$ scalar covariates for this method follows trivially: Consider the first layer in the functional neural network; then, the covariates correspond to the $\ell$-th observation can be seen as the set: $\text{input}_\ell$ = $\{x_{1}(t), x_{2}(t), ..., x_{K}(t), z_{1}, z_{2}, ..., z_{J}\}$. \hyperref[fig:1]{Figure 1} provides a schematic overview of the network structure. Then, suppressing the index $\ell$, (this expression does not change with the observation number) the $i$-th neuron of the first hidden layer corresponding to the $\ell$-th observation is formulated as:
\begin{align*}
    v_i^{(1)} = g\left(\sum_{k = 1}^{K}\int_{\mathcal T} \beta_{ik}(t)x_{k}(t)\di t + \sum_{j = 1}^{J}w_{ij}^{(1)}z_{j} + b_i^{(1)}\right),
\end{align*}
where
\begin{align*}
    \beta_{ik}(t) = \sum_{m = 1}^{M}c_{ikm}\phi_{ikm}(t),
\end{align*}
$\phi_{ikm}(t)$ is the basis function and $c_{ikm}$ is the corresponding basis coefficient. In this formation, the astute reader may notice that $M$ is the same across all $K$ functional weights; it could be the case that the user prefers some functional weight to be defined using a different number of basis functions than $M$ say $M_{k}$ -- indeed, this is left as a hyperparameter.

Given the expansions, we can rewrite this formation of a neuron as:
\begin{align*}
    v_i^{(1)} &= g\left(\sum_{k = 1}^{K}\int_{\mathcal T} \sum_{m = 1}^{M_k}c_{ikm}\phi_{ikm}(t)x_{k}(t)\di t + \sum_{j = 1}^{J}w^{(1)}_{ij}z_{j} + b^{(1)}_i\right)\\
    &= g\left(\sum_{k = 1}^{K}\sum_{m = 1}^{M_k}c_{ikm}\int_{\mathcal T} \phi_{ikm}(t)x_{k}(t)\di t + \sum_{j = 1}^{J}w^{(1)}_{ij}z_{j} + b^{(1)}_i\right).
\end{align*}
This form was proven \citep{thind2020deep} to be a universal approximator in the same vain that the conventional neural network is \citep{cybenko1989approximation}. It is important to note that we can linearly transform the domain and the form of the neuron remains unchanged and thus remains a universal approximator. The argument is as follows: take $t = \alpha u + \beta \implies $ $\di t = \alpha \di u$, with appropriate changes in the boundary values of the integral and then define $c'_{ikm} = \alpha c_{ikm}$. This fact is practically relevant because it allows one to linearly transform the domain to another domain that may be more suitable for the task at hand. 

As observed earlier (with the conventional neural network), running through this set of layers will result in a matrix containing class probabilities. We can asses performance in a number of ways but for the purposes of this article, we consider primarily the proportion of correct classifications details of which are given in \hyperref[sec:3]{Section 3}.

We can now turn our attention to the optimization process of such models. We consider the usual backpropogation algorithm \citep{rumelhart1985learning}. Given our generalization and reworking of the parameters in the network, we see that the set $\theta$ of the parameters in the network is:
\begin{align*}
    \theta = \bigg\{\bigcup_{k = 1}^{K} \bigcup_{m = 1}^{M_{k}} \bigcup_{i = 1}^{n_{1}} c_{ikm}, \bigcup_{u = 1}^{U} \bigcup_{j = 1}^{J_{u}} \bigcup_{i = 1}^{n_{u}} w_{iju}, \bigcup_{u = 1}^{U} \bigcup_{i = 1}^{n_{u}} b_{iu}\bigg\}.  
\end{align*}
This set exists for every observation, $\ell$. We are trying to optimize for the entirety of the training set, so we will move slowly in the direction of the gradient. The rate at which we move, which is called the learning rate, will be denoted by $\gamma$. The specific update for some parameter $a \in \theta$ is: $ a = a - \gamma \Bar{a}$ \citep{ruder2016overview} where $\Bar{a}$ is the derivative averaged across a number of observations. A more in-depth and detailed explanation can be found in \cite{ESL}.

We would also like to emphasize that there are far fewer parameters found in the functional neural networks presented here in comparison with conventional methods. Consider a longitudinal data set where we have $N$ observations and $P$ scalar repeat measurements of some covariate at different points along its continuum. Observe that the number of parameters in the first layer will be $(P + 1)\cdot n_1$ for the usual neural network; however, in our network, the number of parameters in the first layer is a function of the number of basis functions $M$, we use to define the functional weight -- this number, as a consequence of good practice, will be less than $P$; there is no need to have a functional weight that interpolates across all our observed points -- we prefer a smooth effect across the continuum to avoid fitting to noise. Therefore, the number of parameters in the first layer of our network is $(M + 1)\cdot n_1$ where $M < P$.

At the forefront of the black-box reputation of neural network is the inordinate number of weights and biases. It can be nearly impossible to make sense of why these weights move in the direction that they do given that they all move in some conjunction with one another -- the cause and effect is unclear. Unlike the conventional neural network, the method presented here estimates functional weights that are practically similar to the functional regression model. In the case of multiple neurons, we take the average of the estimated functional weight $\hat{\beta}_{k}(t) = \sum_{i=1}^{n_1}\hat{\beta}_{ik}(t)/{n_1}$. Over iterations of the network, as it is trained, we can see movement of the functional weight over its domain. Since these parameters can be visualized, it can be much easier to garner intuition about the relationship between the functional covariates and the response. Our estimates of these will be presented in the examples to come.

Lastly, we would like to discuss the large assortment of hyperparameters that are available to us in the modelling process and their selections. A full list of these can be found in the \hyperref[sec:appendix]{Appendix}. Due to the large number of hyperparameters, there can be a large amount of variation in final outputs. As a result, it is important to consider appropriate tuning paradigms to maximize the performance of the method for a particular context. We implement grid search approaches through the use of \texttt{tfruns}, an \textbf{R} package that allows users to tune neural networks. In the development process, we built our model on top of the \texttt{keras} library which is conveniently compatible with \texttt{tfruns}. The process is simple -- we define a set of possible values for each hyperparameter and then various combinations of these choices are trained and tested for; the model with the best score as indicated by some measure (e.g. correct classification) is chosen to be the final model. From here, we can implement this model for the actual comparisons where there will be different splits of the data so as to reduce bias. This general approach is applied for all models used in the comparisons to come; effort is made to ensure that every model is given similar treatment so as to increase the validity of the final comparisons.

\section{Real Data Applications} \label{sec:3}
\subsection{Modelling Paradigm}
\label{sec:3.1}
Before diving into modelling for a number of data sets and comparing their relative performance across numerous methods, we define the measures we will use throughout the rest of this article. As alluded to earlier, we primarily concern ourselves with correct classification, however, we also provide information on sensitivity, specificity, the positive predictive value, and the negative predictive value. Moreover, we provide uncertainty measures in the form of the variation associated with the classification error.

For completion, we will provide definitions of these measures. We begin with true positive rate sensitivity and specificity. These are also known as the true positive rate (TPR) and true negative rate (TNR). These measures describe how accurately the classifiers correctly label a class. Sensitivity looks at the rate at which the classifier labels the class as its true label, and specificity looks at the rate at which the classifier assigns a false label as false. For the non-binary classification problems we encounter in this work, we report a single sensitivity and specificity value. These values are computed as the average of all the sensitivities and specificities that exist in the confusion matrix. In the end, we report the cross-validated versions of these averages in our results.

Another set of measures we report are the cross-validated positive and negative predictive values (PPV and NPV, respectively). The PPV is computed as the ratio of the number of true positives and the total number of positive labels assigned. Similarly, the NPV is computed as the ratio of the total number of true negatives and total number of negative labels assigned. For non-binary classification problems we generalize in a similar manner described in the previous paragraph.

The variation of the classification error arises because we utilize cross-validation to measure performance between different classifiers. The K-fold cross-validated mean squared prediction error is defined as:
\begin{align*}
    \text{MSPE} = \sum_{k=1}^K \sum_{l \in S_k}^N\left(\hat{y}_{l}^{(-k)} - y_{l}\right)^2/N_{k},
\end{align*}
 where $S_k$ is the $k$-th partition of the data set, $N$ is the number of observations, in the particular fold, and $\hat{y}_{l}^{(-k)}, l \in S_k,$ is the predicted value for $y_l$ by training the functional neural network using the rest of the $K-1$ partitions of the data set. The number of data points in $S_k$ depends on the number of folds. For the real world examples however, we report the accuracy which is defined as follows
 \begin{align*}
    \text{Accuracy} = 1 - \text{MSPE}
 \end{align*}
Additionally, we report the standard deviation on the accuracy. This is computed by looking at the accuracy of each fold and computing the standard deviation on this set of accuracies. 

Some of the models we will use to compare with functional neural networks include the functional linear model using both a basis and an FPCA approach \citep{cardot1999functional}, a non-parametric functional linear model \citep{ferraty2006nonparametric}, and a functional partial least squares model \citep{preda2007pls}. The FPCA and partial least squares methods include variations that apply a penalization penalty to the second derivative in the process of estimating the functional weight, $\beta(t)$ \citep{hall2006properties, kramer2008penalized}. The multivariate methods that will compared include random forests \citep{breiman2001random}, gradient boosted trees \citep{friedman2001greedy}, support vector machines \citep{cortes1995support}, and conventional neural networks.

Some comments on tuning: an effort is made to tune the models presented using the appropriate tools for that particular method. For example, with respect to random forests, we tune the number of nodes and variables chosen within each cross-validated run. Once the best model is picked, then we use it to make predictions on the test (or holdout) set. That is to say, we do not allow for information leakage between our training and test sets but at the same time, provide ample opportunity for this model to find the best parameters for the given context. Much of the tuning for the methods is done through pre-built functions such as those in the \texttt{fda.usc} \citep{fda.usc} package. With respect to neural networks, beyond the grid search, we provide even further advantage in our simulations, details of which are provided in \hyperref[sec:4]{Section~\ref{sec:4}}.

\subsection{Wine Data}

Here, we consider a data set built from the infrared spectrum of wines. These data are publicly available by Professor Marc Meurens, Universit Catholique de Louvain (\url{https://mlg.info.ucl.ac.be/index.php?page=DataBases}). It consists of $n = 123$ observations of the mean infrared spectrum on $256$ evenly-spaced wavelengths and alcohol content (independently measured). Each row is a tuple $(\boldsymbol{S}_i(\boldsymbol{\lambda}), y_i)$ representing the spectrum and class label, respectively. Note, here $\boldsymbol{\lambda}$ denotes the vector of wavelengths that are sampled during the spectroscopy measurement. The class labels are constructed explicitly as follows,
\begin{equation}
y_{i} \in\{1,2\} \text { with }\left\{
\begin{array}{l}
1 \longleftrightarrow \text { greater than 12\% } \\
2 \longleftrightarrow \text { less than or equal to 12\%} \\
\end{array}\right\}
\end{equation}
which corresponds with \citep{dai2017optimal}. The percentages refer to the alcohol content by volume.
In this example, we considered both functional and multivariate methods. The functional methods include the usual functional linear model, and the FPCA, functional non-parametric approaches, and functional partial least squares alluded to earlier in \hyperref[sec:3.1]{Section 3.1}. With respect to multivariate methods, we considered random forests (RF), gradient boosting approaches (GBM), and support vector machines (SVM). Lastly, we also consider the conventional neural network -- a comparison that is explored further in simulations presented later in the article. 

With respect to the multivariate methods, the data were treated as expected -- the discrete time points along the domain were treated as separate variables. In this case, we had 256 such points \textit{i.e.} 256 covariates for each of these multivariate models. For the functional methods, we first transformed the observations into their functional forms using 49 Fourier basis functions. We used one functional covariate -- the functional observation obtained using the raw data. The results are obtained using a $5$-fold cross-validation.

\hyperref[table:winepred]{Table 1} displays the 5-fold cross-validated accuracy, its standard deviation, sensitivity, specificity, positive predicted value, negative predicted value for thirteen functional and multivariate models referenced in \hyperref[sec:3.1]{Section 3.1}. The best model for this example, according to the cross-validated accuracy was the functional neural network. A close second was the gradient boosted decision trees and the functional partial least squares model with the derivative penalization both of which differed at the third significant digit. 

\begin{center}
\setlength{\tabcolsep}{5pt}
\setlength{\columnseprule}{0.4pt}
\scalebox{0.56}{
\label{table:winepred}
 \begin{tabular}{c | c | c | c | c | c | c } 
 \hline
 \texttt{Methods} & \texttt{CV Accuracy} & \texttt{CV Sensitivity} & \texttt{CV Specificity} & \texttt{CV PPV} & \texttt{CV NPV} & \texttt{SD Accuracy} \\ [1ex] 
 \hline
  Functional Linear Model & 0.87 & 0.74 & 0.96 & 0.92 & 0.86 & 0.044 \\ [1ex] 
  \hline
  Functional Non-Parametric & 0.79 & 0.81 & 0.76 & 0.67 & 0.88 & 0.044 \\ [1ex]
  \hline
  Functional PC Regression & 0.88 & 0.81 & 0.93 & 0.87 & 0.89 & 0.056 \\ [1ex] 
  \hline
  Functional PC Regression (Deriv Penalization) & 0.87 & 0.81 & 0.91 & 0.85 & 0.88 & 0.058 \\ [1ex] 
  \hline
  Functional PC Regression (Ridge Regression) & 0.88 & 0.81 & 0.93 & 0.87 & 0.89 & 0.056 \\ [1ex] 
  \hline
  Functional PLS & 0.88 & 0.81 & 0.94 & 0.88 & 0.88 & 0.032 \\ [1ex] 
  \hline
  Functional PLS (Deriv Penalization) & 0.92 & 0.87 & 0.95 & 0.90 & 0.92 & 0.048 \\ [1ex] 
  \hline
  Support Vector Machine & 0.91 & 0.81 & 0.96 & 0.92 & 0.90 & 0.052 \\ [1ex] 
  \hline
  Neural Network & 0.80 & 0.52 & 0.97 & 0.93 & 0.78 & 0.11 \\ [1ex] 
  \hline
  Generalized Linear Model & 0.53  & 0.56  & 0.52 & 0.40 & 0.65 & 0.14 \\ [1ex] 
  \hline
  Random Forest & 0.90 & 0.86 & 0.92 & 0.85 & 0.93 & 0.077\\ [1ex] 
  \hline
  Boosted Trees & 0.92 & 0.94 & 0.91 & 0.85 & 0.96 & 0.039 \\ [1ex] 
  \hline
  Functional Neural Network & 0.92 & 0.91 & 0.94 & 0.88 & 0.94 & 0.059 \\ [1ex] 
 \hline
\end{tabular}}
\captionof{table}{Results of the Wine data set including the 5-fold cross-validated accuracy, its standard deviation, sensitivity, specificity, positive predicted value, negative predicted value for thirteen functional and multivariate models referenced in \hyperref[sec:3.1]{Section 3.1}. Functional neural networks performed the best with respect to accuracy in this example.}
\end{center}



\subsection{Fungi Data set}
Fungal infections in humans impose a considerable disease burden on the world population. Some examples of fungal infections are tuberculosis, meningitis, and fungal asthma. Techniques in molecular biology aid in the study of these fungi. These analyses lend themselves to generating functional data. In this section, we use functional data created by Lu, Sha et al., sourced by William Fonzi, and edited by Hoang Anh Dau~\citep{Lu2017}. \\

The data (\href{http://www.timeseriesclassification.com/description.php?Dataset=Fungi}{data source}) contains functional observations of the rDNA internal transcribed spacer (ITS) region of 51 strains of 18 fungal species. These species define the classes we want to identify. The functional data is a product of High Resolution Melt (HRM) analysis, which is a technique used in molecular biology for the detection of mutations, polymorphisms, and epigenetic differences in DNA samples. It requires precisely ramping up temperature of a sample containing rDNA and then measuring the corresponding melting (dissociation) curve of the sample. Therefore, the continuum of these functional data is temperature. Melt data within the ITS region in the rDNA of a particular fungal species remains consistent among different strains, but differs for different species -- which is a key feature of the analysis and helps in the classification task.

To summarize, this data set has $n=204$ functional observations $\left(\boldsymbol{F}_i, y_i\right)$, where ${i=1,\ldots,n}$ and $\boldsymbol{F}_i$ corresponds to the HRM functional data for a particular temperature value and $y_i$ corresponds to the fungal class membership. There are 18 different classes, with 201 equally spaced temperature steps (in degrees Celcius). They are labelled 1 to 18 based on the species.

We are particularly concerned with how various functional methods handle a problem with multiple classes. The models compared are the same functional models we have analyzed previously. Since there are 18 classes with a fairly small sample size, we were forced to use a 2-fold cross-validation for the comparison. However, even with this small sample size, we observe a noticeable difference in predictive ability amongst the models. We build the functional observations using 39 basis functions. The sequence of points we consider are the temperature points from 75$^{\circ}$C to 100$^{\circ}$C.

\hyperref[table:fungiPred]{Table 2} shows the cross-validated accuracy, its standard deviation, sensitivity, specificity,  positive predicted value, negative predicted value for five functional models referenced in \hyperref[sec:3.1]{Section 3.1}. While keeping in mind the the small sample size, we can observe immediately that the functional neural network outperforms all other models. All of the models seem to perform better than purely guessing, but we observe considerable improvement in performance from the functional neural network. The FNN has nearly a 30\% improvement in accuracy compared to the next best model. There was no need to compare the generalized functional linear model here (seeing as there are eighteen classes) but we note that the usual functional linear model and the functional non-parametric approach seem to perform second best. We also note that there are relatively low standard deviations of these models. Most of the problems for all models seem to occur with sensitivity as they all perform well with respect to specificity.

\begin{center}
\setlength{\tabcolsep}{5pt}
\setlength{\columnseprule}{0.4pt}
\scalebox{0.56}{
\label{table:fungiPred}
 \begin{tabular}{c | c | c | c | c | c | c } 
 \hline
 \texttt{Methods} & \texttt{CV Accuracy} & \texttt{CV Sensitivity} & \texttt{CV Specificity} & \texttt{CV PPV} & \texttt{CV NPV} & \texttt{SD Accuracy} \\ [1ex] 
 \hline
  Functional Linear Model & 0.41 & 0.43 & 0.96 & 0.45 & 0.96 & 0.048 \\ [1ex] 
  \hline
  Functional Non-Parametric & 0.38 & 0.37 & 0.96 & 0.44 & 0.96 & 0.020 \\ [1ex]
  \hline
  Functional PC Regression & 0.17 & 0.17 & 0.95 & 0.21 & 0.95 & 0.041 \\ [1ex] 
  \hline
  Functional PC Regression (Deriv Penalization) & 0.13 & 0.14 & 0.94 & 0.20 & 0.94 & 0.10 \\ [1ex] 
  \hline
  Functional Neural Network & 0.78 & 0.75 & 0.98 & 0.80 & 0.98 & 0.062 \\ [1ex] 
 \hline
\end{tabular}}
\captionof{table}{Fungi data set results of the cross-validated accuracy, its standard deviation, sensitivity, specificity,  positive predicted value, negative predicted value for five functional models referenced in \hyperref[sec:3.1]{Section 3.1}. In this example, the functional neural networks performed the best in classifying the 18 classes.}
\end{center}



\subsection{Orange Juice Data}
Quantitative tabulation of sugar content in fruit juices is crucial for food industries involved in commercial-scale manufacturing and for the regulatory bodies that watch over them. Here we consider another data set made available by Professor Marc Meurens, Universit Catholique de Louvain (\url{https://mlg.info.ucl.ac.be/index.php?page=DataBases}).  This data set is constructed from the infrared spectrum of orange juice and the task is to determine the saccharide level: ``high" or ``low". It consists of $n = 218$ observations of the mean infrared spectrum on $700$ evenly-spaced wavelengths and saccharide level. Each row is a tuple $(\boldsymbol{S}_i(\boldsymbol{\lambda}), y_i)$, where ${i=1,\ldots,n}$ and $\boldsymbol{S}_i$ represents the measured spectrum. Note, here $\boldsymbol{\lambda}$ denotes the vector of wavelengths that are sampled during the spectroscopy measurement. The class labels are constructed explicitly as follows,
\begin{equation}
y_{i} \in\{0,1\} \text { with }\left\{
\begin{array}{l}
0 \longleftrightarrow \text { greater than 40\% } \\
1 \longleftrightarrow \text { less than or equal to 40\% } \\
\end{array}\right\}
\end{equation} 
As was the case in the Wine data example, we compared both functional and multivariate methods in this comparison. There 700 points along the continuum which we used to develop the functional observations. This also implies that the multivariate methods used 700 covariates in the model-building process. A 65-term Fourier basis expansion was used to build the functional observations. Lastly, we use a 5-fold cross-validation to compare results. 

\begin{center}
\setlength{\tabcolsep}{5pt}
\setlength{\columnseprule}{0.4pt}
\scalebox{0.56}{
\label{table:oj}
 \begin{tabular}{c | c | c | c | c | c | c } 
 \hline
 \texttt{Methods} & \texttt{CV Accuracy} & \texttt{CV Sensitivity} & \texttt{CV Specificity} & \texttt{CV PPV} & \texttt{CV NPV} & \texttt{SD Accuracy} \\ [1ex] 
 \hline
  Functional Linear Model & 0.81 & 0.82 & 0.82 & 0.84 & 0.77 & 0.062 \\ [1ex] 
  \hline
  Functional Non-Parametric & 0.69 & 0.70 & 0.84 & 0.84 & 0.73 & 0.052 \\ [1ex]
  \hline
  Functional PC Regression & 0.79 & 0.76 & 0.84 & 0.84 & 0.73 & 0.052 \\ [1ex] 
  \hline
  Functional PC Regression (Deriv Penalization) & 0.66 & 0.74 & 0.59 & 0.70 & 0.63 & 0.093 \\ [1ex] 
  \hline
  Functional PC Regression (Ridge Regression) & 0.76 & 0.72 & 0.82 & 0.83 & 0.69 & 0.027 \\ [1ex] 
  \hline
  Functional PLS & 0.73 & 0.74 & 0.72 & 0.77 & 0.68 & 0.037 \\ [1ex] 
  \hline
  Functional PLS (Deriv Penalization) & 0.67 & 0.74 & 0.62 & 0.71 & 0.64 & 0.092 \\ [1ex] 
  \hline
  Support Vector Machine & 0.63 & 0.58 & 0.75 & 0.76 & 0.58 & 0.038 \\ [1ex] 
  \hline
  Neural Network & 0.58 & 0.90 & 0.23 & 0.62 & 0.43 & 0.085 \\ [1ex] 
  \hline
  Logistic Regression (GLM) & 0.51 & 0.50 & 0.53 & 0.58 & 0.45 & 0.056 \\ [1ex] 
  \hline
  Random Forest & 0.72 & 0.70 & 0.76 & 0.79 & 0.65 & 0.024\\ [1ex] 
  \hline
  Boosted Trees (GBM) & 0.71 & 0.71 & 0.72 & 0.76 & 0.65 & 0.048 \\ [1ex] 
  \hline
  Functional Logistic Reg. (fGLM) & 0.49 & 0.62 & 0.34 & 0.55 & 0.40 & 0.086 \\ [1ex] 
  \hline
  Functional Neural Network & 0.81 & 0.77 & 0.85 & 0.86 & 0.77 & 0.083 \\ [1ex] 
 \hline
\end{tabular}}
\captionof{table}{Orange juice data set results of the cross-validated accuracy, its standard deviation, sensitivity, specificity,  positive predicted value, negative predicted value for fourteen multivariate and functional models referenced in \hyperref[sec:3.1]{Section 3.1}. The functional linear model and the functional neural network approaches performed the best.}
\end{center}

\hyperref[table:oj]{Table 3} displays the cross-validated accuracy, its standard deviation, sensitivity, specificity,  positive predicted value, negative predicted value for fourteen multivariate and functional models referenced in \hyperref[sec:3.1]{Section 3.1}. Most models seemed to perform well relative to just guessing, with the functional linear model and the functional neural network performing the best. While these two have very similar classification accuracy, we note that the functional neural network has a $2\%$ greater standard deviation in the accuracy. The key difference between the two seemed to be their sensitivity performance. Many of the multivariate methods did not perform well in comparison with the functional methods.



\subsection{Phoneme Data}
Phonemes can be understood as a sound or group of sounds percieved to have the same function for a given spoken language or dialect. The English phoneme /k/ is an example: cat, kitten, cougar. Until recently, understanding and categorizing phonemes has been of interest to linguisticians only, but with the recent rise in computation power, the field of speech recognition received a large boost in predictive accuracy and power. 

Here we consider a phoneme data set sourced from \url{https://rdrr.io/cran/fda.usc/man/phoneme.html}. This data set is a subset of a larger dataset which was first used by Trevor Hastie, Andreas Buja, and Robert Tibshirani~\citep{Hastie1995}, and then later used in the ``Elements of statistical learning" textbook. The textbook describes that the data originated from the TIMIT (Acoustic-Phoenetic Continuous Speech Corpus, NTIS, US Dept. of Commerce) and comments that it is a widely used resource for research in speech recognition. The dataset was formed such that there are five phoneme classes: ``sh" as in ``she", ``dcl" as in
``dark", ``iy" as the vowel in ``she", ``aa" as the vowel in ``dark", and
``ao" as the first vowel in ``water". The data was collected by asking 50 male (self-identified) speakers to say particular words that emphasize the different phoneme classes and then recording the audio time-series speech data. Originally the data is time-series, but the functional observations in the data set are the log-periodogram and thus the domain is frequency. The periodiogram can be seen as the auto-correlation of a time-series signal, or the magnitude-squared of its Fourier transform. It tells you how much of the ``signal power" is located in a particular frequency (practically, one would consider the discrete analogue of this).

To summarize, this data set has $n=2000$ functional observations $\left(\boldsymbol{S}_i, y_i\right)$, where ${i=1,\ldots,n}$ and $\boldsymbol{S}_i$ corresponds to the log-periodograms and $y_i$ corresponds to the phoneme class membership. There are 5 different classes, with 400 observations in each class, and thus the classes are balanced. The classes ``sh", ``iy", ``dcl", ``aa", ``ao", are labelled 1 to 5, respectively.

As with the Wine example, we considered both functional and multivariate methods. The functional methods are the same as previous with the exception of the generalized functional linear model which will not handle multiple classes. With respect to multivariate methods, we considered the same methods as well including the conventional neural networks.

For the multivariate methods, the discrete values associated with the continuum points were input. In this case, we had 150 such points i.e. 150 covariates for each of these (multivariate) models. For the functional methods, we first transformed the observations into their functional forms using 65 Fourier basis functions. In this particular example, we used three functional covariates -- namely, the phoneme curves along with their first and second derivatives. The results are obtained using a 10-fold cross-validation.

\begin{center}
\setlength{\tabcolsep}{5pt}
\setlength{\columnseprule}{0.4pt}
\scalebox{0.56}{
\label{table:phoneme}
 \begin{tabular}{c | c | c | c | c | c | c} 
 \hline
 \texttt{Methods} & \texttt{CV Accuracy} & \texttt{CV Sensitivity} & \texttt{CV Specificity} & \texttt{CV PPV} & \texttt{CV NPV} & \texttt{SD Accuracy} \\ [1ex] 
 \hline
  Functional Linear Model & 0.79 & 0.79 & 0.94 & 0.80 & 0.95 & 0.05 \\ [1ex] 
  \hline
  Functional Non-Parametric & 0.22 & 0.22 & 0.80 & 0.25 & 0.80 & 0.053 \\ [1ex]
  \hline
  Functional PC Regression & 0.78 & 0.78 & 0.94 & 0.79 & 0.94 & 0.07 \\ [1ex] 
  \hline
  Functional PC Regression (Deriv Penalization) & 0.66 & 0.66 & 0.91 & 0.67 & 0.91 & 0.077 \\ [1ex] 
  \hline
  Functional PC Regression (Ridge Regression) & 0.78 & 0.78 & 0.94 & 0.80 & 0.94 & 0.06 \\ [1ex] 
  \hline
  Functional PLS & 0.79 & 0.79 & 0.94 & 0.80 & 0.94 & 0.059 \\ [1ex] 
  \hline
  Functional PLS (Deriv Penalization) & 0.76 & 0.76 & 0.94 & 0.77 & 0.94 & 0.65 \\ [1ex] 
  \hline
  Support Vector Machine & 0.91 & 0.91 & 0.98 & 0.91 & 0.98 & 0.032 \\ [1ex] 
  \hline
  Neural Network & 0.85 & 0.85 & 0.96 & 0.89 & 0.96 & 0.16 \\ [1ex] 
  \hline
  Random Forest & 0.90 & 0.90 & 0.97 & 0.90 & 0.97 & 0.029 \\ [1ex] 
  \hline
  Boosted Trees & 0.86 & 0.86 & 0.96 & 0.87 & 0.96 & 0.047 \\ [1ex] 
  \hline
  Functional Neural Network & 0.91 & 0.91 & 0.97 & 0.92 & 0.97 & 0.043 \\ [1ex] 
 \hline
\end{tabular}}
\captionof{table}{Classification results for the Phoneme data set including information about the cross-validated accuracy, its standard deviation, sensitivity, specificity,  positive predicted value, negative predicted value for twelve functional and multivariate models referenced in \hyperref[sec:3.1]{Section 3.1}. The five classes were separated best by support vector machines and functional neural networks.}
\end{center}

\hyperref[table:phoneme]{Table 4} shows the classification results for the Phoneme data set including information about the cross-validated accuracy, its standard deviation, sensitivity, specificity,  positive predicted value, negative predicted value for twelve functional and multivariate models referenced in \hyperref[sec:3.1]{Section 3.1}. We observe that the functional neural networks is competitive with the best supper vector machine classifiers. 
Notably, the conventional neural networks perform well relative to other functional methods, but does not perform as well as the functional neural networks. 

\section{Simulation Studies} \label{sec:4}

The purpose of these simulations is to test the predictive ability of functional neural networks relative to other methods in a more controlled setting. We make an effort to tune all the models used in the comparison. The models we consider are: functional linear models, conventional neural networks, and functional neural networks. Tuning for the functional linear model involved a grid search to find the optimal constraint penalty $\lambda$ for the functional weight. With respect to the two neural network models, we used a grid search to help find the best combination of hyperparameters. As an additional boost to the aid the performance of the conventional neural network, we initialized weights multiple times and picked the set that seemed to have the local optima for the generated data set.

These simulations consist of 150 replicates each for which we generate 300 functional observations. We use the FunData \citep{JSSv093i05} package to generate these observations. The package uses the truncated Karhunen-Loeve representation: 
\begin{align*}
    f_{i}(t)=\sum_{m=1}^{M} \xi_{i, m} \phi_{m}(t)
\end{align*}
Where the set of values $\xi_{i, m}$ is simulated using a normal distribution with a zero mean. This set of values serves as the scores of an approach that also goes by functional principal component analysis. The $M$ orthonormal basis functions are calculated based on a pre-specified basis system e.g. Fourier basis. More details can be found in \cite{JSSv093i05}. Additionally, we add noise to each observation $i$, in the form of $\epsilon_{i}$ which arises from the standard uniform distribution The functional observations are made using 35 Fourier basis functions and we use only the functional data associated with the raw data for the simulation scenarios. Our train/test paradigm consisted of splitting the data generated in \textit{each} replicate into a training and test set -- each of which contained 150 observations. Models were built on the training set and then predictions were made using the covariates of the test set. These predicted classes $\hat{y}$ were compared with the given label, $y$. 

We concern ourselves with three different simulation scenarios. The first consists of generating functional observations from different basis functions. In the first, we generate functional observations from either the Fourier or Polynomial basis depending on whether or not a realization from a uniform random variable is greater than 0.5 or not. We label curves generated using the Fourier basis as class 0 and the Polynomial basis curves as class 1. The models have these curves inputted in the appropriate manner and their errors, sensitivity, specificity, PPV, and NPV are recorded. The second scenario consists of similar generations of data i.e., from the Fourier and Polynomial basis but are then non-linearly transformed using a sinusoidal curve. The purpose was to asses how well these curves perform when the data is transformed in some non-linear way. This can be difficult problem to classify due to the short interval of the domain. In the final scenario, we generate sinusoidal curves for both classes. The only difference is with respect to the amplitude of the curves. Class 1 has an amplitude that is 60\% of class 2.

We use boxplots to measure the error in each simulation replicate. This error is the same as the MSPE defined earlier in the article. The general framework is: for each simulation replication, we will retrieve the test error for each model as recovered using the previously described paradigm. 

\begin{figure}[h!]
    \centering
    \includegraphics[scale = 0.4]{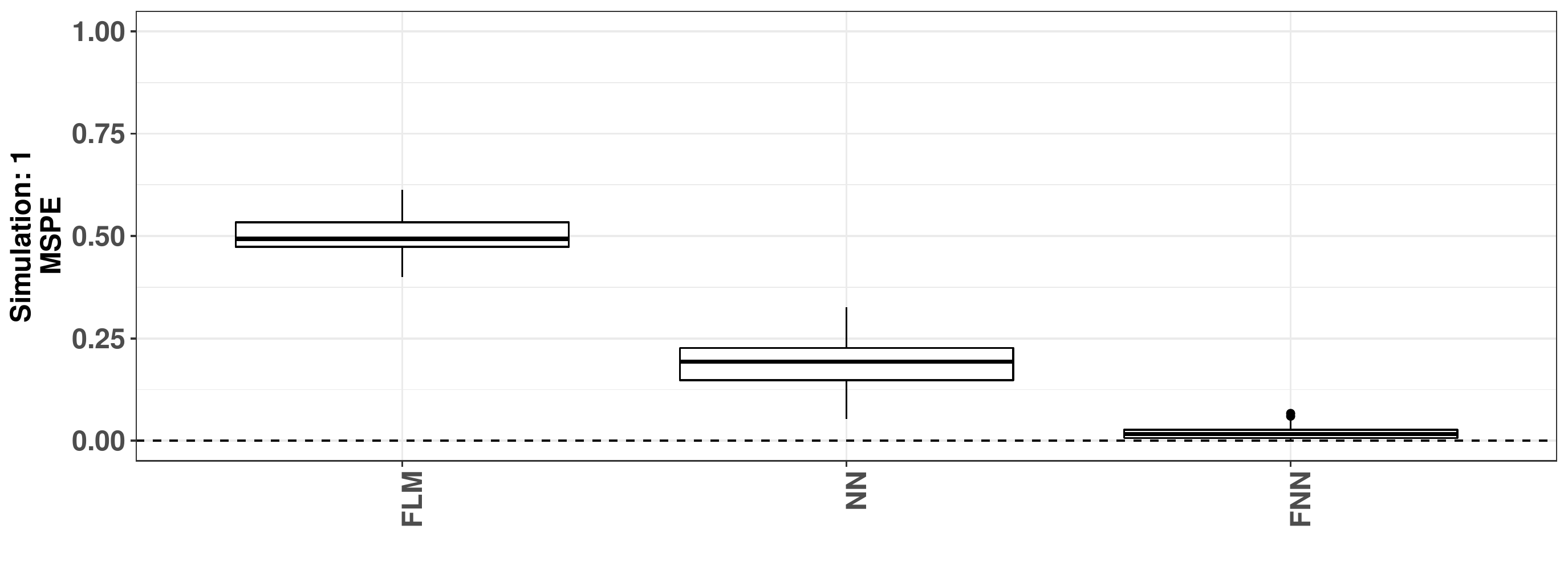}
    \caption{Boxplot comparing the mean squared prediction errors (MSPE) values of the functional linear model (FLM), conventional neural network (NN) and functional neural networks (FNN) obtained after running simulations associated with the first simulation scenario.}
    \label{fig:sim1}
\end{figure}

\begin{figure}[h!]
    \centering
    \includegraphics[scale = 0.4]{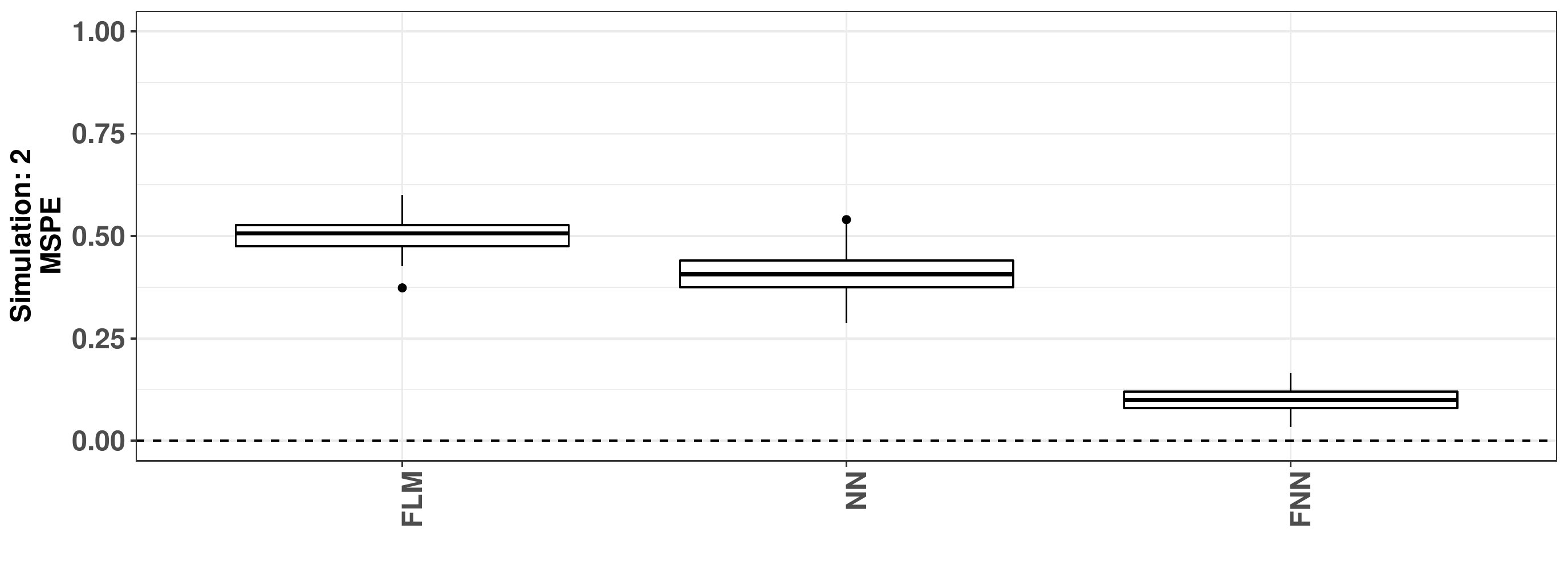}
    \caption{Boxplot comparing the mean squared prediction errors (MSPE) values of the functional linear model (FLM), conventional neural network (NN) and functional neural networks (FNN) obtained after running simulations associated with the second simulation scenario.}
    \label{fig:sim2}
\end{figure}

In the first two simulation scenarios, the functional neural network clearly outperforms the other two methods. The functional linear model, while performing well in a number of the real world applications, seems to finish a distant third here, not predicting better than random selection. In the first simulation scenario the conventional neural network does notably better than the functional linear model but falls short to the functional neural network which had an average accuracy of nearly 98\%; these results can be observed in \hyperref[fig:sim1]{Figure 2} in the form of the MSPE. In the second simulation scenario, the regular neural network does much closer to the functional linear model albeit still an improvement but again, pales in comparison with the functional neural network -- these results are given in \hyperref[fig:sim2]{Figure 3.}

\begin{figure}[h!]
    \centering
    \includegraphics[scale = 0.4]{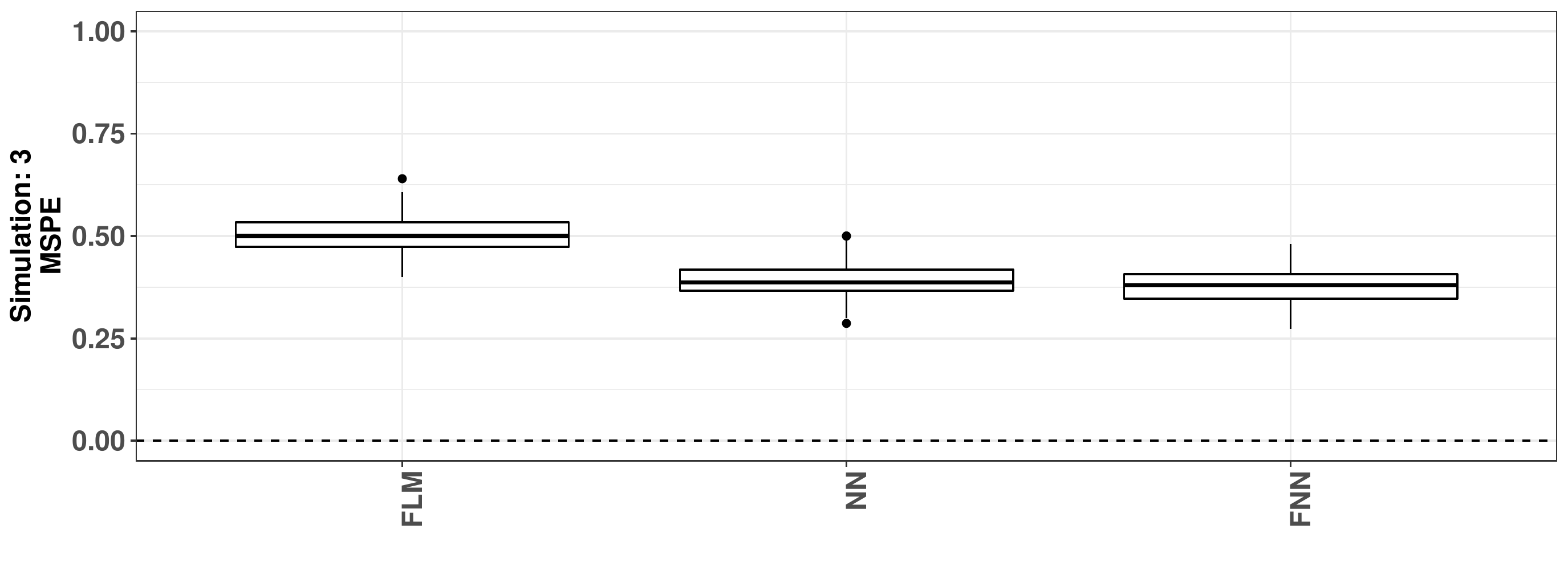}
    \caption{Boxplot comparing the mean squared prediction errors (MSPE) values of the functional linear model (FLM), conventional neural network (NN) and functional neural networks (FNN) obtained after running simulations associated with the third simulation scenario.}
    \label{fig:sim3}
\end{figure}

The third simulation scenario seems to pose much more of a problem for all the models as evident in \hyperref[fig:sim3]{Figure 4}. While the functional neural network did indeed perform the best again, we note that it is much closer to the conventional neural network than in the two previous scenarios. 
This has to do with the nature of this simulation -- since the curves are very similar, and because of the addition of random error, it can be quite difficult for any model to differentiate between these curves; the two distinct classes are generating curves from essentially the same source when error is taken into consideration. Despite this, both the functional neural network and conventional neural network perform better than simply guessing indicating merit. Moreover, we remind the reader that the conventional neural network was given an additional layer of tuning by allowing for optimal selection of starting weights. This luxury was not provided to the functional neural network but, despite this, still outperformed regular networks. \hyperref[table:five]{Table 5} contains the values for the cross-validated error, sensitivity, specificity, PPV, NPV, and standard deviation from these simulations.

\begin{center}
\setlength{\tabcolsep}{5pt}
\setlength{\columnseprule}{0.4pt}
\scalebox{0.56}{
\label{table:five}
 \begin{tabular}{r | c | c | c | c | c | c } 
 \hline
 \texttt{Model} & \texttt{CV Error} & \texttt{CV Sensitivity} & \texttt{CV Specificity} & \texttt{CV PPV} & \texttt{CV NPV} & \texttt{SD Error} \\ [1ex] 
  \hline
  Functional Linear Model - Simulation 1 & 0.50 & 0.49 & 0.50 & 0.51 & 0.51 & 0.041 \\ [1ex] 
  \hline
  Neural Network - Simulation 1 & 0.18 & 0.82  & 0.80 & 0.80  & 0.82  & 0.052  \\ [1ex] 
  \hline
  Functional Neural Network - Simulation 1 & 0.019  & 0.99  & 0.96  & 0.96  & 0.99 & 0.014  \\ [1ex] 
  \hline
  Functional Linear Model - Simulation 2 & 0.50 & 0.50 & 0.49 & 0.48 & 0.51 & 0.038 \\ [1ex] 
  \hline
  Neural Network - Simulation 2 & 0.40 & 0.59 & 0.59 & 0.59 & 0.60 & 0.049 \\ [1ex] 
  \hline
  Functional Neural Network - Simulation 2 & 0.10 & 0.91 & 0.88 & 0.88 & 0.91 & 0.29 \\ [1ex] 
 \hline
  Functional Linear Model - Simulation 3 & 0.50 & 0.49 & 0.50 & 0.50 & 0.48 & 0.043  \\ [1ex] 
  \hline
  Neural Network - Simulation 3 & 0.39 & 0.47  & 0.74 & 0.65 & 0.59  & 0.039  \\ [1ex] 
  \hline
  Functional Neural Network - Simulation 3 & 0.37  & 0.51  & 0.73  & 0.65  & 0.60  & 0.039 \\ [1ex] 
 \hline
\end{tabular}}
\captionof{table}{Resulting values of the cross-validated accuracy, its standard deviation, sensitivity, specificity,  positive predicted value, negative predicted value for all three simulation scenarios.}
\end{center}

\section{Conclusions} \label{sec:5}

In this article, we discussed classification tasks and presented how the proposed class of models performs relative to other predictive approaches. This functional neural networks incorporate a methodology that is at the cross-section of functional data analysis and deep neural networks. We highlighted advantages of such an approach and discussed various aspects of the methodology including optimization and associated hyperparameters.

We showed the versatility of functional neural networks by applying them in a number of classification scenarios. This includes application in data sets with multiple classes and implementation in a number of popular data sets such as the Wine and Orange Juice data. We further cemented the predictive ability of our method through a number of simulation scenarios where it was pitted against conventional neural networks and the functional linear model. Furthermore, we laid the groundwork for why the functional parameters discussed in the article have a more interpretable nature to them than the scalar parameters traditionally estimated in conventional neural networks.

There is ample room to further extend the methodology presented here. In fact, this is just the first in what can become a family of deep learning methods. A natural next step is to consider the case when the response is a curve rather than a scalar value or a class. This work is underway and involves a clever adjustment of the function-on-function FDA method. Another extension relates to shrinkage methods being used in conjunction with the estimation of the functional weights of the network.

\section*{Supplementary Materials}

Data and R Codes used for each application and simulation study presented in this paper are available at
\texttt{https://github.com/caojiguo/FunClassifiers}

\section*{Acknowledgments}
 This research was supported by the Natural Sciences and Engineering Research Council of Canada (NSERC) 
Discovery grant (RGPIN-2018-06008), the NSERC Strategic Projects - Group 494291-2016-STPGP, and the Canada Research Chairs program.
\bibliography{mybib}

\section*{Appendix - Model Parameters}
\label{sec:appendix}
\begin{center}
\setlength{\tabcolsep}{15pt}
\setlength{\columnseprule}{0.2pt}
\scalebox{0.65}{
\label{table:one}
 \begin{tabular}{c | c | c } 
 \hline
 \texttt{Parameter} & \texttt{Type} & \texttt{Details} \\ [1ex] 
 \hline
  $\beta(t)$ & Estimated & Coefficient function found by the FNN. \\ [1ex] 
  \hline
  $w$ & Estimated & The scalar covariate weights. \\ [1ex] 
 \hline
  $b$ & Estimated & The intercept in each neuron. \\ [1ex] 
 \hline
  Number of lyers & Hyperparameter & The depth of the FNN. \\ [1ex] 
 \hline
  Neurons per layer & Hyperparameter  & Number of neurons in each layer of the FNN. \\ [1ex] 
 \hline
  Learn rate & Hyperparameter & The learning rate of the FNN. \\ [1ex] 
 \hline
  Decay rate & Hyperparameter & A weight on the learning process across training iterations for the FNN. \\ [1ex] 
  \hline
  Validation split & Hyperparameter & The split of training set during learning process. \\ [1ex] 
  \hline
  Functional weight basis & Hyperparameter & The size of $M$ for the estimation of $\beta(t)$. \\ [1ex] 
  \hline
  Training iterations & Hyperparameter & The number of learning iterations. \\ [1ex] 
  \hline
  Batch size & Hyperparameter & Subset of data per pass of the FNN. \\ [1ex] 
  \hline
  Activations & Hyperparameter & The choice of $g(\cdot)$ for each layer. \\ [1ex] 
  \hline
  Early stopping & Hyperparameter & Stops the model building process if no improvement in error. \\ [1ex]
  \hline
  Dropout & Hyperparameter & Randomly drops some specified percentage of neurons from one layer to the next. \\ [1ex]
  \hline
\end{tabular}}
\captionof{table}{A list of the parameters in the network.}
\end{center}

\end{document}